\documentclass[manuscript,screen]{acmart}

\AtBeginDocument{%
  }

\setcopyright{none}
\copyrightyear{2023}
\acmYear{2018}
\acmDOI{XXXXXXX.XXXXXXX}

\acmConference[Conference acronym 'XX]{Make sure to enter the correct
  conference title from your rights confirmation emai}{June 03--05,
  2018}{Woodstock, NY}
\acmPrice{15.00}
\acmISBN{978-1-4503-XXXX-X/18/06}

\usepackage{colortbl}
\usepackage{graphicx} 
\usepackage{url}
\usepackage{stfloats}
\usepackage{tabularx}

\begin{document}

\title [Equitable AI Research Roundtable] {The Equitable AI Research Roundtable (EARR): Towards Community-Based Decision Making in Responsible AI Development}

\author{Jamila Smith-Loud}
\email{jsmithloud@google.com}
\orcid{1234-5678-9012}
\author{Andrew Smart}
\author{Darlene Neal}
\author{Amber Ebinama}
\author{Eric Corbett}
\author{Paul Nicholas}
\author{Qazi Rashid}
\author{Anne Peckham}
\author{Sarah Murphy-Gray}
\affiliation{%
  \institution{Google Research}
  \streetaddress{One Market}
  \city{San Francisco}
  \state{CA}
  \country{USA}
}

\author{Nicole Morris}
\affiliation{%
 \institution{Emory University School of Law, Innovation and Legal Tech}
 \country{USA}}

\author{Elisha Smith Arrillaga}
\affiliation{%
  \institution{Student Success Solutions}
  \country{USA}}

\author{Nicole-Marie Cotton}
\affiliation{%
  \institution{UC Berkeley, Center for Information Technology Research in the Interest of Society (CITRIS)}
  \country{USA}}

\author{Emnet Almedom}
\author{Olivia Araiza}
\affiliation{%
  \institution{UC Berkeley, The Othering and Belonging Institute}
  \country{USA}}

\author{Eliza McCullough}
\affiliation{%
  \institution{Partnership on AI}
  \country{USA}}

\author{Abbie Langston}
    \affiliation{%
     \institution{PolicyLink}
\country{USA}}

\author{Christopher Nellum}
\affiliation{%
 \institution{The Education Trust-West}
  \country{USA}}

\renewcommand{\shortauthors}{Smith-Loud et al}

\begin{abstract}
  This paper reports on our initial evaluation of The Equitable AI Research Roundtable–a coalition of experts in law, education, community engagement, social justice, and technology. EARR was created in collaboration between a large tech firm, nonprofits, NGO research institutions, and universities to provide critical research based perspectives and feedback on technology's emergent ethical and social harms. Through semi-structured workshops and discussions within the large tech firm, EARR has provided critical perspectives and feedback on how to conceptualize equity and vulnerability as they relate to AI technology. We outline three principles in practice of how EARR has operated thus far that are especially relevant to the concerns of the FAccT community: how EARR expands the scope of expertise in AI development, how it fosters opportunities for epistemic curiosity and responsibility, and that it creates a space for mutual learning. This paper serves as both an analysis and translation of lessons learned through this engagement approach, and the possibilities for future research. 
\end{abstract}

\maketitle

\section{Introduction}
It is by now well established that the risks and harms of AI technology are disproportionately distributed towards groups that are already structurally vulnerable \cite{eubanks2018automating, benjamin2020race, o2017weapons, obermeyer2019dissecting}. Recognition of this issue among some AI researchers, industry, and governments has led to the development of Responsible AI, a loose set of industry and research governance practices which attempts to address the social harms associated with machine learning \cite{raji2020closing, schiff2020principles, mittelstadt2019principles}. The problems for the nascent field of Responsible AI are to understand the source of this risk to the public, its root cause, and to either prevent or mitigate the impacts on those who are most affected. To do so, Responsible AI needs to develop methodologies and processes for incorporating the viewpoints and perspectives of people who are not traditionally at the table during the development process. 

A major challenge in this area is incorporating diverse forms of expertise in relevant disciplines on these complex social issues, which fall outside the expertise of computer science \cite{mulligan2019thing}. Most research and industry practice on Responsible AI and social impacts of AI technology are carried out by those institutions who are closely associated with the technology being evaluated, and who are, intentionally or unintentionally, sympathetic to it and to those who implement it. This narrow epistemic lens from which AI development is conducted contributes to a whole host of downstream social impacts. In this paper, we present our evaluation of The Equitable AI Research Roundtable (EARR) –a novel organizational intervention – which was created to center the voices of experts from historically marginalized groups.

EARR is a community-based AI discussion group created within a large tech company by a team of socio-technical researchers with backgrounds spanning from nonprofit business, disability advocacy, human rights law, computer science, anthropology, economic development, to education advocacy. It generally functions as a coalition consisting of external nonprofit and research organization leaders who are equity experts in the fields of education, law, social justice, AI ethics, and economic development and the tech company’s socio-technical research team; coming together, their goals are to: (1) center the voices of historically marginalized groups, (2) qualitatively understand and identify harm type and severity within AI models and technologies that majorly impact these groups, and (3) provide actionable feedback and validation of potential harms from the perspectives of the most vulnerable and suggestions on mitigations/control mechanisms that could make the AI model or product safer and fairer for all. The coalition meets monthly to review, discuss, and provide expert equity and harm guidance on particular AI technologies or research ideas that teams across the company present to the group. This feedback outputs in the form of discussion summaries, adversarial datasets, user control and transparency enhancements, and/or fairness or harm strategy reframing and inputs into team’s fine tuning of their AI model and UI, user outreach strategy, and product fairness approach. 

EARR offers a novel approach to AI governance that addresses the gaps in technical tools, principles, standards and regulations. EARR additionally presents a novel form of community-based feedback as a forum for the exchange of equity and fairness ideas, AI ethics conflict resolution, and harm identification. In our initial analysis of EARR, we highlight three functions of EARR that are especially relevant and valuable for the FAccT community: how it expands the scope of expertise in AI development, how it fosters opportunities for epistemic responsibility, and that it creates a space for mutual learning. 

We describe these three functions and discuss the advantages and disadvantages of the EARR approach within big tech companies, how it compliments and expands others approaches in the field of Responsible AI, and how the EARR approach can be improved and implemented in other environments.

\section{Related Work}
\subsection{Current Landscape of Responsible AI}
The rapid pace of AI development has eclipsed a rigorous and responsible understanding of the social and ethical risks associated with this technology. Throughout history, nascent technology has often gotten ahead of its scientific underpinnings, ethical understanding and engineering knowledge in order to launch rapidly to keep pace with market trends; this leads to new technology falling behind in scientific, ethical, and engineering best practices \cite{leveson2016engineering}. As evidence of bias, disproportionate negative impacts and social harm on structurally vulnerable people from machine learning began to accumulate in the last decade, technology firms, civil society, and academia began to publish responsible and ethical AI principles meant to guide how firms develop and deploy AI technology responsibly \cite{jobin2019global}.  

The field of ML fairness began to focus on mathematical and algorithmic ways to define, measure and address unrepresentative datasets and biased outputs \cite{kleinberg2018algorithmic}. This narrow focus on quantification, measurement, technical tools and algorithmic fixes for these issues failed to capture the larger social structural and sociotechnical root causes of algorithmic harm \cite{selbst2019fairness}. The assumption in the technology industry often being that the solution to problems caused by technology is more technology and the people who developed the problematic systems are the only people capable of fixing these systems; furthermore, these technical solutions are rarely developed in conversation with those most impacted \cite{passi2019problem, martin2020participatory}. 

Despite the progress that Responsible AI writ large has made in characterizing and mitigating the harmful impacts of computational systems, there remains a large conceptual and practical gap between epistemologies of privilege and power that characterize large technology firms collectively and the lived experiences of people at the sharp end of a machine learning model. These so-called “downstream” impacts on human beings are abstracted into variables and probability distributions \cite{selbst2019fairness}. Technical approaches to mitigating bias without a social science informed understanding prioritize a certain type of knowledge centered around computer science and computational systems. Gebru, following other scholars, has described this as the hierarchy of knowledge in machine learning where technical acumen in the form of coding skills, mathematical and quantitative reasoning are seen as the only legitimate forms of knowledge \cite{GebruTalk}. The hierarchy of knowledge in AI is a form of gatekeeping that prevents other kinds of knowledge, what Foucault called subjugated knowledge \cite{allen2017power}, from entering into conversations and decision making processes inside large tech firms and elite academic institutions that develop AI. 

To begin confronting this hierarchy of knowledge, EARR created an open dialogue format in order to give a forum for external researchers and scholars with domain expertise in relevant areas to be in direct engagement with researchers, product teams, and engineers at a large AI firm. A fundamental goal with EARR is to expand the scope of expertise involved in the ML development lifecycle by involving experts with both the lived experience and specialized academic knowledge to address epistemic gaps. EARR is also intended to be mutually beneficial in that external experts from different disciplines gain at least partial insight into a company’s AI technology development process. 

EARR also differs from traditional user research and external advisory committees in important ways. EARR has been operating since 2019 and thus forms a continuous set of relationships among members, as opposed to one-off studies or engagements. This allows deeper relationships to form among all those participating from the tech firm, NGOs, non-profits, and universities. This model of long-term engagement is intentionally meant to encourage these relationships and to advance equity this requires novel approaches to addressing conflicts, seeking feedback from stakeholders, and centering relationships; which contrasts with the typical focus on producing deliverables, meeting timelines, and to achieving benchmarks. 

\subsection{Context for community engagement and coalitions}
Community engagement in the education, social science and nonprofit research community, is steeped in equal partnership with identified community members who have come together in efforts to address a challenge, need, or tackle a long-standing question. This type of engagement includes meaningful involvement from all participants to ensure the joint research is appropriately implemented and analyzed, and that the community’s expertise is elevated and incorporated into all aspects of the research plan \cite{ahmed2010community}. 

Several models for community engagement exist in research, including participatory research, action-based participatory research, evaluation, and coalition building. According to a 2010 study in the American Journal of Public Health, “Community engagement in research is a process of inclusive participation that supports mutual respect of values, strategies, and actions for authentic partnership of people affiliated with or self-identified by geographic proximity, special interest, or similar situations to address issues affecting the well-being of the community of focus. Community engagement is a core element of any research effort involving communities. It requires academic members to become part of the community and community members to become part of the research team, creating a unique working and learning environment before, during, and after the research.” \cite{ahmed2010community}. 

Utilizing the coalition-style framework in social science and nonprofit organizations, lends itself to emphasizing the core mission and values of problem-solving with innovative practices through research and advocacy efforts in partnership with communities for actionable change. Coalition members come from diverse backgrounds, including demographics, education, workforce sector, and skill set. Participating in a coalition brings shared goals around promoting equity and justice and changing the impact of how various social systems operate within communities. Within a coalition, it is a given that the members possess the knowledge, power, and agency as experts in their field and domain, to move their agenda forward with a focus on their agreed principles and mission. 

Working with research coalitions in the nonprofit and public sector space as a form of community engagement is often not viewed as an essential or imperative ingredient in AI development, or as a technical contribution that can move products and tools closer to launch. Though product teams are intending to build fair and equitable AI as part of their core mission and values, there are limits to understanding the complexity of issues around fairness, equity, and harms. Introducing EARR as a coalition of diverse stakeholders, means that product teams can meaningfully engage with various aspects of community facing considerations directly from experts in the field, moving responsible AI work from theoretical implications to engaging partners in the field on the complex processes and recommendations that stem from particular domain and knowledge expertise. 

\section{Equitable AI Research Roundtable: Context and Origin}

The EARR program takes place at a large technology company that recognized the need to  create a space where external experts in equity where brought together with internal researchers and developers to jointly move further towards the common goal of ensuring AI technology is ethical and fair. Internally many of our teams where working through how to understand fairness and ethics within specific contexts. The goal of EARR was to move these conversations outside of tech specific silo, we believed that there was much to learn from experts who had thought about the hard questions of conceptualize problems and strategies toward fairer outcome, who had measured and tested outcomes and who had context specific expertise. Out of the above interests, we created EARR. 

EARR was initially developed with the goal of promoting a space for knowledge sharing with experts who focus on research and data-based analysis of historically marginalized communities and/or groups.We thought the coalition format would provide a useful and more equitable basis for conversations around who is being impacted by algorithmic systems, how they are being impacted, and alternative ways to think about these issues. The goal of this coalition is to center the voices and experiences of those who are potentially most impacted by harmful deployment of AI, engage diverse perspectives around technology harms, and assist in the conceptualization and understanding of AI/ML  related concerns.

EARR members consist of experts in issues of equity stemming from K-12 education, law, AI and tech ethics, academia, economic justice, research and policy. These fields were chosen because they represent understudied areas in AI research, but also areas where AI is having an increasing impact.  

\subsection{Initial goals and composition of the coalition}
We created EARR with an initial set of community expert criteria: we wanted to collectively encompass a wide spectrum of issue areas, to work across domains on matters of fairness and equity, and to have an organizational focus on research and evidence- driven analysis or assessments. Some of the intended initial outcomes included create a space to discuss research approaches, document concerns as they related to specific communities or context, translate feedback generated in the sessions to product or teams specific guidance documents and to advocate for more intentional community based methods to be used internally. The initial cohort comprises researchers and experts from a wide array of policy groups, research organizations and non-profits. Table ~\ref{table:EARRMembers} summarizes the domains of expertise and initial member roles.

\begin{table*}
  \caption{EARR Domains of Expertise and Initial Member Organizational Roles}
  \label{tab:commands}
  \begin{tabular}{ccl}
    \toprule
    Domain & Position\\
    \midrule
    \text{Education Equity} & NGO Director, Professor of Education \\
    \text{Tech Equity \& Law} & Professor of Law\\
    \text{Math + CS Equity} & Professor of Education\\
    \text{Social, Economic, \& Neighborhood Studies} & Director of Research\\
    \text{Democracy, LGBTQ Citizenship, \& Disability Studies} & Researchers\\
    \text{Economic Development and Health Equity} & Director of Research\\
    \text{Criminal Justice} & Director of Research and Policy\\
    \bottomrule
    \label{table:EARRMembers}
  \end{tabular}
\end{table*}

The first initial consideration in forming EARR were that there should be shared benefit among the partners in the coalition. This involved ensuring that the large tech firm did not define what the benefit and the goal of partnership was for member organizations, and an acknowledgement of internal knowledge gaps and external expertise. 

The second consideration was an intentional goal of creating partnerships (of which this paper is an outcome). We wanted to identify interdisciplinary experts, manage existing relationships with the tech firm. We also wanted to manage expectations in relation to large technology corporations and smaller academic groups or non-profits. In addition to this, we had to manage internal stakeholder expectations at the large tech firm around expectations and impact. 

Thirdly, we had to come to consensus about the scope of the engagements, which involved creating and agreeing on the types of engagement and the focus, which initially stated as research exchanges around the intersection of the domains represented by the coalition of experts. However, as discussed, this shifted over time to prioritize creating space for sharing knowledge and feedback on specific products. 

Finally, we agreed upon a model of compensation which prioritized a fair and appropriate compensation for the time and expertise that EARR members gave. This process is ongoing and interactive to ensure that compensation remains fair and up to date with external market conditions. 

\subsection{How does EARR work?}
The job of the EARR team at the large tech company is to coordinate dialogue among experts in K-12, higher education, nonprofit organizations, and local and government policy organizations that focus on legal systems and advocacy efforts. Within this group we encourage the diversity of professional and lived experience to expand our understanding of the varied impacts of AI across structures and demographics and help fill in areas necessary to mitigate against any harms to particular communities and populations. EARR participants work in multiple capacities within the roundtable, including sharing current and relevant research studies related to their field of study, particular to their organization, and critical for AI development. The members are also introduced to product areas from teams who are exploring broader social equity implications in their development and launch of models and tools, helping to identify impacts and provide context for potential harms and inequities. 

During the first year of EARR we convened roundtable discussions around education equity, racial equity issues in AI and big data, risk assessment frameworks for AI, data cards and dataset documentation, human rights issues and big tech, and a critical race approach to algorithmic fairness \cite{hanna2020towards}. 

Within the latter capacity of EARR, product teams approach internal EARR organizers with questions of how to think about harms to individuals and populations as they engage with their product or tool, which harms should be prioritized for mitigation attempts ‘fixing’, how they should think about use cases differently for each harm and control, and the broader implications to social equity they should be aware of in their development process. Internal EARR organizers then work with product teams to develop their presentations, create discussion questions, and coordinate time with EARR members to engage in a joint session on conceptualizing harms and providing recommendations for designing mitigations. 

Different from traditional UX research, product teams provide EARR members with the background and context of their product/tool, real examples of how the product/tool currently works, and any relevant failures they have already uncovered. Within the framework of responsible AI principles, product teams are looking to EARR to learn from the failures, uncover additional failures and harms, and find ways to incorporate context of those harms into future iterations of their development. EARR was not intended or created to have veto power over decision making in product development, nor is it a formal part of the AI governance processes - although product teams do get referred to EARR as part of this governance process. In the spirit of knowledge sharing, EARR provides inputs to these internal processes. 


\subsection{EARR and social positionality}
EARR grew out of a need to establish a sustainable and strategic process for research-based external engagements with established leaders and researchers who are deeply invested in understanding and mitigating issues around fairness and equity.  As an internal team of researchers and analysts  our job is typically  provide foundational and applied research. and assessments as inputs to the multiple points of decision making throughout the company.  We fulfill our mission and goals of advocating for users, particularly those from communities who have been historically marginalized or are particularly vulnerable to impacts and harms, by ensuring that experts related to these communities interest, needs and experience are an integral part of our research process. In an ideal situation the focus on fairness, equity, and ethical AI would occur at the very beginning of product and research development processes - while teams are still ideating real world issues to solve and defining problems and solutions. 

The research team that organizes EARR are ethics workers at a large technology firm that researches, develops and deploys AI. As such we are acutely aware of our social positionality and how that influences our stance, and we see EARR as an essential counter to our perspectives. Even though many of the company's responsible AI researchers come form civil rights and nonprofit backgrounds, it was crucial for us to select external EARR members that are representative of the social science domains from which themes such as ethics, equity, fairness, bias, and harm derive. Big tech researchers now have the epistemic lens of rapid AI development and innovation so there needs to be a balanced perspective from a societal and anthropological lens. The researchers are united in their goal to work toward organizational change at the large tech firm, specifically to help the development of AI become more sensitive to the perspectives and lived experiences of differently situated people, whose life experiences and academic training gives them lucidity on equity issues the tech company lacks.

\section{Methodology} 
This paper is a both an analysis and a reflection on the leanings that have emerged from the EARR engagements over the last four years. our experiences managing, coordinating, and engaging in research conversation as the EARR roundtable. In this capacity, we observe an evolving exchange in which EARR members and researchers at the large tech firm develop and deepen intellectual relationships. Our approach is post-hoc, collective reflection on our experiences and the outputs of EARR sessions.  

We reflected on our experiences managing EARR. During the course of the four years of EARR qualitative data from bi-monthly EARR sessions are collected in the form of product team presentations, meeting notes including discussion questions and responses, and occasional post-session follow up on particular observations and feedback related to metrics or validation of findings. These artifacts are shared between internal EARR  organizers, EARR members and product teams to promote transparency and collaboration, and support EARR’s contribution to product team’s knowledge base in the short and long term. 

One main output during the beginning stages of EARR’s sessions included a worksheet intended to help tech workers who seek coaching on applying equitable practices to the research and development of equitable AI. It featured guidance from external experts who participated in a year-long partnership. The Roundtable met quarterly to discuss how to improve the research and development of fair, safe, and socially beneficial AI applications that are accountable to people.

In addition to the sessions and its documentation of findings, EARR organizers and product teams often hold follow up meetings to discuss the group’s observations, clarify intended outcomes and findings, develop adversarial datasets to enrich testing of AI models, and cross-validate any themes from the discussions.

\section{Findings: Three Principles in Practice}
Throughout the course of conducting EARR workshops and meetings we have found three salient principles as the approaches to the work that have developed in practice that we would like to highlight.
\subsection{Principle 1: Expand the scope of expertise in AI development}

As the field of Responsible AI has grown over the last several years, there is an increasing imperative toward those responsible goals to expand the conception of expertise. EARR challenges the traditional notions of relevant knowledge to be an AI practitioner and to contribute to the development and understanding of these complicated models. Increasingly regulators and policy makers are demad understanding of AI models relative to sensitive contexts, community impacts and assessments of harm. This questions require expertise out side of traditional computer science and data science backgrounds.

Just as engineers, research scientists, and product managers are experts in technology development, EARR members are experts in their respective fields and bring vast experience and knowledge into conversations about the possible harms and biases of AI on marginalized communities. We have come to understand the unique mission of EARR as a coalition-style and expert-led community engagement series; it is important to then examine its novelty in the technology sector where we combine the equity-based voice of EARR with the technical development and design of AI. For example, the internal EARR team observed that there are certain social science concepts such as redlining and human rights principle and domain areas such as patent law that teams across the tech company was able to gain a better understanding on after roundtable discussions with external experts. Teams have been able to use these concepts to shape expectations on what a "fair" model output should look like.      

As nonprofit and social science sectors continue to embrace the structure and research power of coalition-building, those same approaches still seem novel in large corporate and technology companies. Artificial intelligence and large technology models and tools develop and launch quickly and serve to make different aspects of the human experience easier, more fun and exciting, but also address real-life challenges in health, education, and the environment. “However, many of these systems are developed largely in isolation of the communities they are meant to serve. In the best case, this may lead to applications that are improperly specified or scoped, and are thereby ineffective; in the worst case, it can lead (and has led) to harmful, biased outcomes for marginalized populations.” \cite{black2020call} 

To combat the harmful or biased outcomes, many product areas seek to engage EARR on the fairness and ethical implications around applied AI technologies and discuss top of mind ethical concerns related to the solutions they are building. To do this, translating mental models across EARR and technology development must include an approach that recognizes and values EARR’s essential critical thought regarding the experiences, analysis, and problem solving that comes with coalition-style engagement. In the rapidly evolving regulatory landscape, there is also an increasing need to involve experts in sensitive domains as defined for example by the GDPR \cite{EUdataregulations2018}.  


Initially EARR was focused on an exchange of research and ideas surrounding equity and technology, however as the group progressed all parties became interested in more direct exchanges of perspectives regarding specific technologies. As EARR has moved from a knowledge sharing forum to one in which EARR interacts more directly with product teams at the large technology firm, there emerged interesting changes and learning from the process. The interaction of alternative experiential standpoints and multiperspectivalism, between our research team, EARR members, and product teams, generated mutual learnings that helped expand the way in which data scientists and product managers gather information about how their models will impact and interact with society - and especially structurally vulnerable groups who will feel the impacts of these system first and be potentially disproportionately impacted. 

A key finding from this process was the reframing of fairness questions: from technical to value-based. Throughout the experiences gathered during EARR sessions this shift in perspective helped teams developing ML-driven technology to see new perspectives; however it at times resulted in exposing gaps that would need more time and knowledge to address. 

\subsection{Principle 2: Measure values and judgment versus numbers}
A primary concern of most product teams in the technology industry is to measure and quantify fairness in machine learning applications. As Selbst et al argue concepts in computer science—such as abstraction and quantification —are used to define notions of fairness and discrimination, to produce fairness-aware learning algorithms, and to intervene at different stages of a decision-making pipeline to produce "fair" outcomes. However, these concepts render technical interventions ineffective, inaccurate, and sometimes dangerously misguided when they enter the societal context that surrounds decision-making systems \cite{selbst2019fairness}. As mentioned earlier within the hierarchy of knowledge, abstraction and quantification enables decisions to be made that appeal to “objectivity”, where quantification is assumed to be one and the same as objectivity. But as Porter points out, “The idea of objectivity is a political as well as a scientific one.” \cite{porter1996trust}  

EARR offers an alternative way to discuss judgements of values based on expertise and experience and in conversation about values. We argue that this actually leads to a stronger sense of objectivity, following Patricia Hill-Collins and Sandra Harding, who argue that questions involving  or affecting marginalized groups should begin with the lives and experiences of those marginalized groups \cite{intemann201025, collins2002black}. Members of marginalized groups have experiences that result from their social positionality as "insider-outsiders." They must not only understand the assumptions that constitute the worldviews of the dominant group to navigate the world, but they also have experiences that conflict with this worldview and generate alternative understandings of how the world works \cite{intemann201025}. In the AI industry, often missing from practices around algorithmic accountability is the question of  what and whose values or lived experiences get to have influence over the development of the algorithm in the first place \cite{black2020call}. Within EARR the framing of questions from quantification, measurement and prioritization of fairness issues in Responsible AI are motivated by dialogue and values for understanding equity, understanding knowledge communities and the practice of knowledge sharing.

For example, in pursuing fairness objectives the large tech firm's research and engineering teams often would like to quantify social groups and demographic data in order to enable measurement and testing of product performance across demographic subgroups. While performance metrics for fairness are an important goal, and fairness failures of commercially available AI-driven products are plentiful \cite{raji2019actionable, buolamwini2018gender}, well-intentioned attempts at algorithmic fairness can have effects that may
harm the very populations these measures are meant to protect \cite{raji2020saving, liu2018delayed}. For example, the tension between intersectionality, the understanding of how interlocking systems of
power and oppression give rise to qualitatively different experiences
for individuals holding multiply marginalized identities, and abstract reductionist approaches to group fairness is often underappreciated \cite{raji2020saving}. EARR provided essential perspective on the complexity and nuance required for understanding the social implications of these tensions.  These teams came to EARR seeking validation of various methods of demographic estimation, however through discussion with EARR members it became apparent that the deeper issue was around trust in Big Tech to collect or steward such demographic data in a safe way, given histories of problematic uses of data. As engineers and product managers focus on optimizing quantified metrics, EARR opens up larger avenues of thinking that do not readily occur to technically focused teams that are pressed for time. EARR provided crucial input that at least exposed technical teams to alternative ways of looking at the problem, and explained why optimizing technical performance does not address the underlying trust issue with using data about sensitive social attributes. EARR enabled a deeper discussion of issues that are normally not thought to be germane to technical discussions, but reveals that sociotechnical systems must be thought of as not independent from the social and power relationships into which they are embedded.  


 
\subsection{Principle 3: Create opportunity for mutual learning}
Mutual learning can be defined as a process of information exchange between organizations within a coalition, in the case of EARR, the exchange of knowledge occurs between big tech employees and the external equity experts who represent communities most adversely impacted by AI. According to Scholz, the term was coined in 1997 during a responsible environmental science case study and describes the extent to which scientists and society self-organized to exchange information for the goal of advancing sustainable development \cite{klein2001transdisciplinarity}. During EARR, there is a methodical process to AI ethics discourse that naturally leads to knowledge sharing and building on ideas to reach the common goal set out by the coalition. The key distinction in this discourse is that the speakers at this discussion round table are uniquely representatives of society and AI researchers - typically conversations in this Responsible AI and ML Fairness space includes only the engineers, AI practitioners, and research scientists who are representative of the organization they develop AI models for. 

In a typical EARR session, the large tech firm provide technical, use case, and market insight on AI models. Internal teams showcase early stage AI research and products to participants and provide a platform for external experts to think more critically about their tech equity and AI ethics research. The external experts provide thought knowledge on harm, ethics, and impact from the viewpoint of communities who are most affected by unfair classification, recommendation, or risk prediction outputs of AI systems. Based on the feedback from the experts, internal big tech teams are able to adjust their framing around the utility of a model or product for the benefit of actionable org-wide Responsible AI changes that may include: diversifying datasets to make models more representative of society, pivoting away from problematic use cases to reduce harm, modifying user communication and outreach strategy to build trust, expanding mental models to be more reflective of the diversity in global society, and/or implementing control mechanisms to ensure model safety. 

We describe two examples during EARR sessions where knowledge exchange prevailed and the tech company was able to widen their narrow epistemic lens. In social science literature, equity is understood as "justice" or "quality of being fair." At the tech company, fairness in ML is mainly perceived as a mathematical equation for the technical implementation of reducing biased outputs. In one EARR session, a product team wanted to investigate how an additional feature would impact across various socio-cultural groups. Would the feature have the same output across all groups and can this be measured? The product team was looking for a mechanism to quantify "fair" product outcomes - to supplement the team's framing, the external experts suggested the team think about equity and fairness in terms of justice from a social lens first.  Viewing fairness issues in AI from a social lens helped AI practitioners to focus on the people that are most impacted by problematic outputs. External experts challenged the product team to think critically about why the additional feature, when used the same way across groups, might still result in different outputs and how societal structures caused this imbalance in adverse effects. Without knowledge rooted in anthropology, economics, sociology, etc., product teams will only be able to provide a formulaic solution that does not reach to the crux of society's problems that massive amounts of data, used to fuel AI technology, reflects. 

The topic of severity and harm is a reoccurring theme in the majority of discussions between the external equity experts and tech company AI practitioners. The only way to identify and understand harm and severity of harm is through the viewpoint of those who experience it. A research team building a text2image solution presented preliminary identified use cases and harms to the external EARR members. They were seeking validation for correct observation of harm, how severe the harm is, and identified groups most likely to be affected by the harm. Before validating technology harm, the EARR group first helped the text2image team to understand the type and levels of existing societal harm for historically marginalized people that is now also reflected in well-documented healthcare, job, surveillance, justice system, entertainment AI solutions. To help build AI that is truly fair for all, EARR members' expertise allows for them to systematically frame their mental models towards a societal perspective that is intended to sync with the technical mitigation the tech firm strives to implement.    

\begin{quote}
“[Coalitions] succeed best when members are open to learning from one another in order to guide joint actions—and along the way they may even change their own thinking.” 
-Kanter and Hayirli, Creating High Impact Coalitions \cite{kanter2022creating}
\end{quote}

The two EARR session instances above consummates Kanter and Hayirli's observation of a coalition. This give-and-take learning between big tech and society leads to enhanced thinking and action around building fairer, equitable AI technology and importantly, trust building - thus leading to more roundtable engagements that result in knowledge sharing and immediate use in model or UI enhancements or fairness, harm, and bias reframing. Trust building is a by-product of mutual learning in a coalition setting similar to a participatory design approach. External experts feel empowered to provide their domain expertise and represent communities in a way that highlights their knowledge on not only social issues, but also technological issues that impact society. By providing information on how their thought leadership impacts research scientists and AI practitioners' decisions to think more critically about AI systems before they launch, external experts become more trustworthy of the tech company and thus, they can potentially relay these relationship strides to their surrounding communities - the users that will ultimately engage with the company’s products. 

The space created to learn and actively use shared knowledge in a coalition is a critical component to the success of EARR. It is not by happenstance that EARR was formed to be a hub for sharing information across social science and computer science perspectives. Furthermore, it was vital for external participants to be experts in a variety of domains - so that there would be knowledge sources that are credible, reliable, and available for immediate use. Taking from the diverse backgrounds in nonprofit and advocacy of the big tech team’s founding EARR members, EARR was strategically designed to be rooted in the transdisciplinarity approach which gives platform to mutual learning and promotes trust building and striving towards common goals: (1) amplifying the voices of the most affected groups of AI impacts, (2) identifying and mitigating technology harm, and (3) propelling equity and fairness technical implementation forward.

\section{Discussion} 
This paper has thus far has explained the origins of EARR's development and the practical mechanics of this engagement model. The principles of practice are summations of the applied approaches that have emerged over the course of our engagement. Although we as a research team had initial goals related to the need of convening and learning from experts, the expected outputs of the engagements were admittedly ill-defined. The typical speed and urgency in which tech companies build products and make decisions gave our internal research team some pause and concern about the viability of this type of engagement, as the goal was around knowledge sharing rather than the more common project goals related to product impact. Our initial hypothesis, could there be mutual benefit from engagement between a tech company and  community based and equity centered experts to the collective understanding of responsibility and ethics in the AI development and governance process, is one that we continue to explore and answer (see Table -\ref{table:Topics}). 

\begin{table*}
  \caption{Topic/Issue covered in EARR engagements (research presentations by internal and external participants)}
  \label{tab:commands}
  \begin{tabular}{ccl}
    \toprule
    Topics Covered in EARR engagements\\
    \midrule
    \text{Algorithmic decision making} \\
    \text{Big Data \& Equity} \\
    \text{Tech \& Education Equity} \\
    \text{Human Rights \& AI} \\
    \text{Equity and the Algorithmic-Driven Hiring} \\
    \text{Sensitive domains (criminal legal system, eduction, health)} \\
    \text{Transparency \& Explainability} \\
    \text{Assessing Risk \& Defining vulnerability} \\
    \text{Predictive Algorithms \& Equity} \\
    \text{Measuring Impacts \& Harm (image models; text models; facial recognition models)} \\
    \bottomrule
    \label{table:Topics}
  \end{tabular}
\end{table*}

\subsection{\textbf{Community of Experts}}

EARR has established itself within the large tech firm as an important source of external expertise for Responsible AI. In addition to providing a forum for the ethics researchers and workers at the tech firm to share knowledge, the EARR engagements have resulted in the development of a community. We recognize that the term community is increasingly a contested term with in research communities, specifically in the ethics and fairness discussion related to AI and ML. Over the last few years there has been an increased focus on community based participatory methods in the Responsible AI field, as well as an increased focus on diversity as relevant to the development and outcomes of AI model deployment. 
EARR's long term engagement with experts was intentional and was not developed to capture user experience, which we see and understand as resulting in a different set of insights and outcomes than we hoped to achieve with EARR.  Community for this project is not a group that we are observing or "researching" but rather the causatum of this engagement overtime.

Our working definition of  “community” is intended be used for both complex and specific contexts; and is not intended to be used as a one-dimensional proxy for sensitive characteristics such as race, ethnicity, gender, nationality, income, sexual orientation, ability, and political or religious belief. Defining community in this engagement has been an iterative process. We initial started with a simple set of criteria for involvement: 1.) Collectively encompass a wide spectrum of disciplines/issue areas; 2.) Applied focus on fairness and/or equity; 3.) Organizational/individual skill set  in evidence based analysis or assessments. Overtime and certainly the value of a long-term engagement strategy has been that our definition of community has evolved to include trusted partners who are linked by common goals around advancing equity and a commitment to engaging in complex research and conceptual questions.

\subsection{\textbf{Relative Context }}
\begin{figure}
\begin{center}
\includegraphics[scale=0.40]{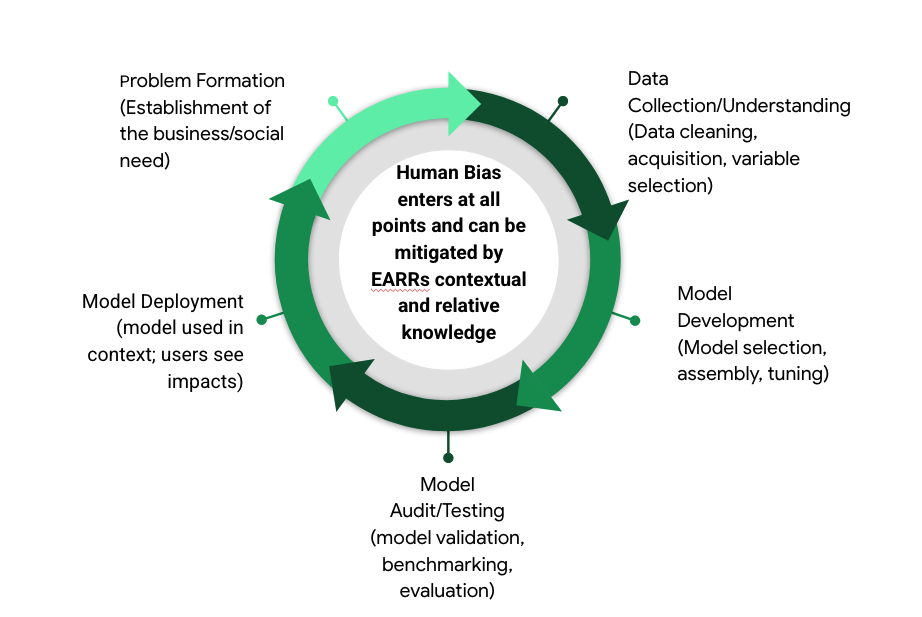}
\end{center}
\caption{The AI development life-cycle and where EARR can influence the conversation.}
\label{fig:lifecycle}
\end{figure}

We have found the value of EARR engagement has been relevant across many types of context and research questions including  the validation of research ideas, localized and domain specific assessments and to adding context specific knowledge to all stages of the model development process (see Figure~\ref{fig:lifecycle}). From the initial problem formation phase to the model deployment phase the domain and interdisciplinary perspective of EARR members have added relative knowledge and context, feedback and critical discussion on relevant interventions. The EARR process has revealed the necessity for understanding of responsible AI issues and areas of concern relative to specific context.


\subsection{\textbf{Equity Lens}}
Fairness in the machine learning context is a contested term. Along with Mulligen et al \cite{mulligan2019thing}, EARR does not argue for a single “proper” conception of fairness, rather an important goal of EARR is to recognize and connect the diversity of ways these terms are defined and operationalized within different disciplinary and professional contexts, and different communities of practice including those of law, social sciences, humanities, and computer science. By fair we mean letting a plurality of perspectives be heard. EARR demonstrates one of the novel instances of community engagement using a coalition of stakeholders with research, advocacy, and equity backgrounds, and big tech partners. 

Responsibility for the social harms of computational systems cannot be properly identified and addressed independently of our social positions within structural processes and networks of social relations, and this recognition is what EARR begins to enable by bringing together the epistemic advantages of centering experts from historically marginalized perspectives. EARR attempts to address epistemic ignorance of the privileged social positions within a large tech firm, as Medina points out, “how to achieve (at least some degree) of meta-lucidity for differently situated subjects, including those in a position of privilege whose life experiences may not have put them in the best position to become lucid about insensitivity” \cite{medina2012epistemology}. 


\section{Conclusion} 
The ongoing productivity of EARR is supported by an active community of experts and stakeholders that share a range of goals, including research and information sharing, and most often come together to engage in discussion and activities that respond to equity-based solutions to technical product development. This type of engagement in the technology community is useful for reaching beyond the capacity of any individual or organization and dives deeper into group expertise that leads to rich discussion, and meaningful and applied recommendations that can prove to protect and uplift the social equity concerns across communities and populations.

\section{Acknowledgments}

The authors would like to thank all the participants in the EARR meetings over the last four years.


\bibliographystyle{ACM-Reference-Format}
\bibliography{EARR_FAccT.bib}


\end{document}